\documentclass[lettersize,journal]{IEEEtran}

\usepackage{amsmath,amssymb,amsfonts}
\usepackage{bbm}
\usepackage{cite}
\usepackage[ruled]{algorithm2e}
\usepackage{graphicx}
\usepackage{textcomp}
\usepackage{xcolor}
\usepackage{subfigure}
\usepackage{hyperref}
\usepackage{multirow}
\usepackage[flushleft]{threeparttable}

\usepackage{amsmath,amsfonts}
\usepackage{algorithmic}
\usepackage{array}
\usepackage{textcomp}
\usepackage{stfloats}
\usepackage{verbatim}
\usepackage{graphicx}
\def\BibTeX{{\rm B\kern-.05em{\sc i\kern-.025em b}\kern-.08em
    T\kern-.1667em\lower.7ex\hbox{E}\kern-.125emX}}
\usepackage{balance}

\newcommand{\com}[1]{
}
\newcommand{\eric}[1]{
}

\newcommand{\diff}[1]{{#1}}  
\newcommand{\diffa}[1]{
    #1
} 
\newcommand{\diffb}[1]{
    #1
} 

\newcommand{\todo}[1]{
} 

\title{\LARGE \bf
Hierarchical Generative Adversarial Imitation Learning with Mid-level Input Generation for Autonomous Driving on Urban Environments
}

\author{Gustavo Claudio Karl Couto$^{*}$ and Eric Aislan Antonelo$^{*}$
\thanks{$^{*}$Gustavo C. K. Couto and Eric A. Antonelo are with the Automation and Systems Engineering Department, Federal University of Santa Catarina,
        Florianópolis, Brazil
        {\tt\small gustavo.karl.couto@posgrad.ufsc.br, eric.antonelo@ufsc.br}.}%
}
\begin{document}

\maketitle

\begin{abstract}
Deriving robust control policies for realistic urban navigation scenarios is not a trivial task. In an end-to-end approach, these policies must map high-dimensional images from the vehicle's cameras to low-level actions such as steering and throttle.
While pure Reinforcement Learning (RL) approaches are based exclusively on \diff{engineered} rewards,
Generative Adversarial Imitation Learning (GAIL) agents learn from expert demonstrations while interacting with the environment, which favors GAIL on tasks for which a reward signal is difficult to derive\diff{, such as autonomous driving}.
\diff{However, training deep networks directly from raw images on RL tasks is known to be unstable and troublesome.
To deal with that, this work
proposes a hierarchical GAIL-based architecture (hGAIL) 
which decouples representation learning from the driving task to solve the autonomous navigation of a vehicle. 
The proposed architecture consists of two modules: a GAN (Generative Adversarial Net) which generates an abstract mid-level input representation,
which is the Bird's-Eye View (BEV) from the surroundings of the vehicle;
and the GAIL which learns to control the vehicle based on the BEV predictions from the GAN as input.
hGAIL is able to learn both the policy and the mid-level representation simultaneously as the agent interacts with the environment.
}
Our experiments \diff{made in the CARLA simulation environment} have shown that GAIL exclusively from cameras (without BEV) fails to even learn the task, while hGAIL, after training \diff{exclusively on one city, was able to autonomously navigate successfully in 98\% of the intersections of a new city not used in training phase.
Videos and code available at: 
\href{https://sites.google.com/view/hgail}{\texttt{https://sites.google.com/view/hgail}}.
}
\end{abstract}


\section{Introduction}
Commonly, Autonomous Driving (AD) has been implemented using individual modules for perception, planning and control organized in a pipeline \cite{juniordarpa, env_perception,surveypipeline,Parizotto2021}.
However, learning approaches have been on the rise, in an attempt to tackle the complexities of AD in different scenarios, in simulation or even in the real world. 
Most of the approaches are based on Behavior Cloning (BC), which uses supervised learning on a set of expert demonstrations collected offline \cite{nvidiabc, videobc, chauffernet, codevilla2018endtoend, codevilla2019exploring}, e.g., with a human driver generating a set of input (observations) and corresponding desired output pairs. The latter approach suffers from covariate shift \cite{efficientReductions, 9156703, ross2011reduction}, since it can not teach robustly the learning agent a trajectory which does not accumulate errors.

Reinforcement Learning (RL) approaches to AD can learn policies that do not present this covariate shift issue, since the agent is able to learn in interaction with the environment, considering the whole sample trajectories and not only independent observation-action samples as in BC. 
However, RL requires the definition of a reward signal, which can be cumbersome to do it considering the complexity of a driver's behavior and its environment. Although Inverse RL can be used for imitation learning purposes 
\cite{Ng2000}, 
it is expensive to run since it executes RL in a loop. Learning in IRL is computationally more expensive than just learning a policy directly from expert demonstrations.

On the other hand, Generative Adversarial Imitation Learning (GAIL) \cite{Ho2016} provides a way to train agents in interaction with their environments, directly from expert demonstrations. This approach has been validated in the CARLA realistic simulator for autonomous driving in urban scenarios previously \cite{gail_carla}, showing it can scale to large environments.
However, in \cite{gail_carla} only fixed routes were considered. Although the agent's architecture was general enough for dynamic routes, with inputs such as the high-level command and the next point of the sparse trajectory in the vehicle's reference frame, the network had limitations for learning a general policy for dynamic routes, i.e., those that can change on the fly (turn left, right, or go straight at an intersection) from the perspective of the agent.

\diff{
In \cite{resnet_rl}, the authors show that training complex mathematical functions such as deep neural networks for online policy learning based on complex input patterns from raw camera images
is unstable. This is because the reward signal is not informative enough for training such large models.
One of the strategies employed by \cite{resnet_rl} to circumvent this limitation is to decouple feature representation from policy learning.
In our context, the decoupling is achieved by learning a mid-level representation
of the vehicle's raw camera images.
For instance, Bird's-Eye View (BEV) representations of the road ahead of the vehicle have been used as mid-level input to trajectory generation and motor control networks in \cite{chauffernet} and \cite{roach}, respectively.
In BEV representation, the scene from the camera is projected onto a top-down view, and regions of interest in the image, such as vehicle lanes and sidewalks, are segmented into different colors \cite{env_perception}.
Using BEV as input to the control policy, the idea is to make policy learning focus on the navigation problem, since it will use a more abstract and simple view of the road ahead.
}

One of the advantages of such approach is enabling transfer to real-world by a relatively easy process: it requires mapping the real-world images to the same abstract representation used to train the agent in simulation. Besides, training agents with mid-level visual inputs such as BEV and others (e.g., optical flow, depth, semantic segmentation, and albedo) make policies learn faster, generalize better and achieve higher task performance \cite{Sax2019}. 
Other works employ semantic segmentation for semantic driving
\cite{Muller2018,Mousavian2019,Yang2018}, offering supporting evidence that mid-level inputs to agents are useful for realistic downstream active tasks \cite{Sax2019}.
So far, BEV has been employed for autonomous driving in urban scenarios in \cite{roach} and \cite{chauffernet}. However, they generate BEV input by using an algorithm which has to access a known map of the city. 
In contrast, in our work, we consider a mid-level \diff{predicted} BEV input, feeding the agent's policy, \diff{that is generated by a separate network that learns simultaneously with the policy network in a hierarchical GAIL (hGAIL) architecture}.
\diff{The proposed hGAIL consists of two modules: a learning module for mapping camera images to BEV representations; and a policy learning module based on GAIL that controls the vehicle by using the predicted BEV as mid-level input.
The first module uses Conditional Generative Adversarial Nets (GANs) \cite{pix2pix} and U-nets to map the images obtained from three frontal vehicle's cameras to the mid-level BEV input. 
The latter module, based on GAIL, outputs steer, acceleration and break signals to drive the vehicle based on the GAN's output.}
Thus, although our agent produces \diff a mid-level representation, it still pertains to the class of end-to-end models, \diff{since the raw camera images are fed as inputs to the hGAIL architecture}.
\diff{Additionally, the} GAIL module's cost function is augmented with a behavior cloning loss \cite{Jena2020} in order to stabilize the policy learning.
Both GAN and GAIL networks learn simultaneously, though with their own cost functions, while the agent interacts with the CARLA simulation environment for urban driving. This approach ensures that both the policy and representation networks are trained using on-policy data, learning from the agent's mistakes.

This work contributes by:
\begin{enumerate}
    \item \diff{
    overcoming the difficulties of policy learning directly from raw pixels
    for the GAIL agent in autonomous driving. 
    This is achieved by decoupling representation learning from the driving task, through learning a Bird’s-Eye View (BEV) representation of the input, which can be processed more efficiently by GAIL.
    }
    \item proposing an end-to-end hierarchical architecture based on both GAN and GAIL (hGAIL) for autonomous urban driving. 
    \diff{The online nature of GAIL allows the agent to experience out-of-distribution samples not existing in the expert dataset, different from Behavior Cloning approaches \cite{h_birdview}.
    After training, the hGAIL agent only has access to a sparse trajectory, being able to predict a more refined desired trajectory that is used by the policy to directly output the steering and throttle motor actuators, unlike approaches that output the trajectory and use a nonadaptive controller to move the vehicle \cite{ral_gan_intention, chauffernet}.
    }
    \item extending previous work  that is applicable to fixed routes \cite{gail_carla} to a more skilled agent able to follow dynamic routes, making the vehicle able to change the route on the fly. \diffa{In addition, this work also tackles navigation in cities with pedestrians and other vehicles, and respecting traffic lights, unlike
    \cite{gail_carla}.
    }
    \item \diff{proposing a GAN network for BEV mid-level representation generation of three channels (desired route, drivable area, and lane boundaries) based on inputs from the raw vehicle's frontal camera images, sparse trajectory and high-level command. 
    This approach enables simultaneous representation learning with policy learning in an online manner, without degrading the GAIL training process.
    }
\end{enumerate}

\section{Related Works}

\subsection{Behavior Cloning (BC)}
The authors of works \cite{codevilla2018endtoend} and \cite{codevilla2019exploring} utilized Behavior Cloning (BC) for Autonomous Driving (AD) in the CARLA simulator.
A large dataset of human driving was collected and augmented using image processing techniques to train end-to-end policies conditioned on the desired route. Further advancements in BC were made in \cite{ral_prob_bc}, which used a large deep ResNet network for feature extraction and fused data from camera, LiDAR, and radar to generate feature maps. 

\subsection{Reinforcement Learning (RL)}

\diffa{In \cite{ral_racing}, an RL agent was trained and tested in a 2D simulator, CARLA, and a real-world car racing task, highlighting the potential use of auxiliary models to address the weakness of RL in needing large amounts of environment interactions.
%
In \cite{roach}, RL is employed}
with a customized reward function to train an agent based on the mid-level BEV representation as input, during the first training phase. 
Afterwards, using the trained agent as an online expert, they trained a second agent through apprenticeship learning, but now in an end-to-end approach with input directly from the vehicle's cameras.
\diffa{Their method evaluated using CARLA surpassed} the default benchmarks trained by BC.

\subsection{GAIL}
The authors in \cite{Kuefler2017} used Generative Adversarial Imitation Learning (GAIL) with simple affordance-style features as inputs to reduce cascading errors in behavior-cloned policies and make them more robust to perturbations. Raw LiDAR readings and simple road features, such as speed and lane center offset, were mapped to turn-rate and acceleration to model human highway driving behavior in a realistic highway simulator. The experiment successfully reproduced human driver behavior while reducing the risk of collisions. 

\diff{In \cite{gail_av}, model-based GAIL (MGAIL)
was used with a large-scale expert dataset (100,000 miles) from San Francisco city, and applied to the task of dense urban self-driving. The discriminator and generator in their MGAIL model consists of Transformer networks, which combine different types of inputs: roadgraph points, traffic light signals, and other objects' trajectories.
They introduce an hierarchical model
which integrates a high-level graph-based search (to generate the intended trajectory) with a low-level transformer-based MGAIL policy. 
Besides, they do not learn mid-level input representations from raw camera images as hGAIL in this work, but use a pretrained independent robotic perception system to generate input features.
Additionally, our controller in hGAIL learns with expert samples generated in the CARLA simulator corresponding to only 36 minutes (8 km) of driving.  
In theory, transferring the trained policy in hGAIL to a real-world setting would require only replacing the BEV network by another network trained on real images.}

\subsection{BEV Mid-level representation}

Reinforcement learning enables an agent to be trained in a closed-loop fashion, resulting in more robust agents. However, this robustness comes with the cost of instabilities that prevent the use of very deep networks, such as ResNet \cite{resnet}. In \cite{resnet_rl}, the authors aim to decouple representation learning from the training of reinforcement learning as a solution to train agents that can process high-dimensional raw inputs, such as images from cameras.

In \cite{ral_gan_intention}, \diff{a hierarchical driving model is proposed, consisting of a driving intention module and trajectory generation module.
The former learns to generate the future trajectory with a GAN, whose generator} is fed with an image from a monocular camera and a local map, cropped from an offline map using the GPS position. 
The intention map is combined with LiDAR data to generate a potential map, which is their mid-level input representation (similar to BEV), subsequently fed to a controller module trained 
to imitate a set of demonstration trajectories. 
\diff{While their BEV prediction has only one channel with the desired trajectory (intention), our BEV in hGAIL produces also other two channels for drivable area and lane boundaries. However, in our approach, we do not combine LiDAR data with the BEV prediction. The final output of their model corresponds to a trajectory function that is used by a nonadaptive controller to drive the vehicle, whereas hGAIL outputs directly the values for the steering and throttle motor actuators.
The training of their networks are offline, opposed to our online training of hGAIL, which is more challenging with respect to the learning task of directly mapping raw frontal cameras images to motor actuators.
}
They evaluate their method on CARLA and also on a real vehicle.

A similar modular approach is explored in \cite{h_birdview} \diff{in the CARLA simulation environment}, which combines a predicted Bird's-Eye View representation
with a pretrained trajectory of 0.5-meter resolution to serve as inputs to a policy network trained with Behavior Cloning. 
\diff{To generate the BEV, first, the raw image is segmentated using a pretrained segmentation network; then, its output} is transformed by an
U-net into an BEV top-down perspective. 
\diff{
It is worth noting that our BEV generation in hGAIL uses a
Conditional GAN that maps directly from raw camera images to the final BEV prediction, while \cite{h_birdview} uses a segmentation network as an intermediate step.
Additionally, our BEV prediction includes the desired trajectory as a BEV channel, while the BEV prediction in \cite{h_birdview} does not include this information, needing access to the points of a dense trajectory as inputs to the policy. 
In contrast, after training, our hGAIL agent only access a sparse trajectory of 50-meter resolution on average, which is 100 times more sparse than the one used in their work. 
Besides, 
they train the BEV network offline in a supervised way and before the policy network is trained by conventional Behavior Cloning, also offline. 
Notice that, in our work, both BEV and policy networks learn online in an integrated way simultaneously and in an adversarial way. Thus, covariate shift situations are better handled with hGAIL, since the policy can learn with samples not included in the expert dataset.
}

\diff{The BEV input representation in hGAIL is inspired by \cite{chauffernet}, which trains a system to autonomously drive a vehicle on CARLA using a Bird's-Eye View representation as input.}
This BEV representation, like our own, contains a road and route representation from a top-down, agent-centered perspective. 
\diff{However, after training, hGAIL does not need anymore to access the city map to generate the BEV of the vehicle, as this is done by a GAN network, unlike in \cite{chauffernet}. 
Based on this BEV input, in their work, a neural network is trained offline to output the trajectory to be followed by a nonadaptive controller. 
They also employ data augmentation and auxiliary losses in this training. 
In contrast to this,
our hGAIL agent learns both the policy and the BEV representation online and simulatenously, i.e., in a closed-loop training process, mapping raw frontal camera images direcly into motor actuator values (steering, and throttle). Thus, the controller also has to be learned in hGAIL, unlike \cite{chauffernet} which learns only the trajectory offline.}

The work in \cite{Muller2018} uses the output of a scene segmentation network as a mid-level input representation.
This enables the transfer of the agent's policy to the real world, since a similar network can be used to generate the mid-level input from images of real scenes.

Our work is the first instance, to our knowledge, where a GAIL is trained using a mid-level input representation generated by a learning module to tackle the intricate task of autonomous driving navigation. Previous works also show that policies with mid-level representations as input can be trained on a simulator, and subsequently transferred to the real world. This is a cost-effective and safe way to perform closed-loop training of agents (since infractions during training of the agent occur in simulation only), in particular, by the GAIL method which provides an interactive learning based on expert demonstrations.

\section{Methods}

\subsection{Conditional Generative Adversarial Networks - CGAN}

CGANs are composed of two neural networks, a discriminator D and a generator G. The function of G in a CGAN \cite{pix2pix} is to translate an image $x$ into an image $y$ by mapping both $x$ and a random noise vector $z$ into an output image $G:\{x,z\} -> y$.
Both D and C seek to optimize the same objective function:

\begin{equation}
\mathop{\mathbb{E}}_{x,y}[\log(D(x,y))] + \mathop{\mathbb{E}}_{x,z}[\log(1-D(x,G(x,z))],
\label{eq:cgan}
\end{equation}
where G tries to minimize it, while D seeks to maximize it. 

Additionally, a L1 distance loss function is added to the final objective, making the generator network G also learn from the true label $y$ as it would happen in a supervised learning task, 
\diff{modeling low-frequency characteristics of images}. 
\diff{The resulting method, PatchGAN \cite{pix2pix}, divides the image into multiple patches to be classified by the discriminator individually and, thus, requires fewer parameters.}

The CGAN is used in our work to generate the Bird's-Eye View (BEV) image representation from the agent's sensors such as frontal cameras and GPS, to be detailed later.

\subsection{Generative Adversarial Imitation Learning - GAIL}

Mathematically, GAIL
\footnote{
\diffa{More details on GAIL are given in Appendix \ref{apndx:gail}.}
}
finds a saddle point $(\pi,D)$ of the expression:
\begin{equation}
\mathop{\mathbb{E}}_{\pi_E}[D(s,a)] - \mathop{\mathbb{E}}_\pi[D(s,a)] - \lambda H(\pi) - \lambda_2 L_{gp}, 
\label{eq:wgail00}
\end{equation}
where $D: S \times A \rightarrow (0,1) $, $S$ is the state space, $A$ is the action space; $\pi_E$ is the expert policy;
\diffa{$\pi$ is the agent's policy};
$H(\pi) $ is a policy regularizer controlled by $\lambda >= 0$ \cite{Bloem2014}.
\diff{The discriminator D will try to increase (\ref{eq:wgail00}), while $\pi$ seeks to minimize it; and $L_{gp}$ is a loss that penalizes the gradient's norm,
constraining the discriminator network to the 1-Lipschitz function space, according to \cite{gulrajani2017improved}.}
The above equation is the Wasserstein loss, used
to alleviate vanishing gradient and mode collapse problems, using
the Wasserstein distance \cite{gulrajani2017improved} between the policy 
distribution and expert distribution, as also done in \cite{Zhang_2020,li2017infogail}.

\subsection{Bird's-Eye View (BEV) representation}
\begin{figure*}[thpb]
  \centering
  \subfigure[Frontal Cameras]{
\label{fig:cameras}  
  {\includegraphics[scale=0.46]{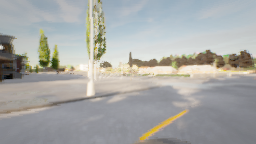}}
  {\includegraphics[scale=0.46]{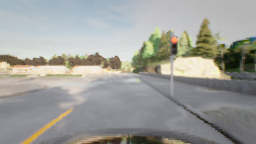}}
  {\includegraphics[scale=0.46]{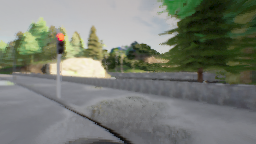}}
  }
  \subfigure[Sparse trajectory]{
  {\includegraphics[scale=0.4]{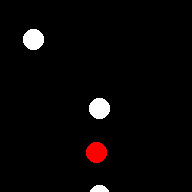}}
  \label{fig:traj_plot} 
  }
  \subfigure[Bird's-Eye View representation (BEV) channels]{
  {\includegraphics[scale=0.4]{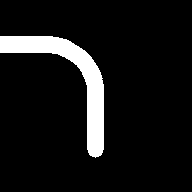}}
  \quad
  {\includegraphics[scale=0.4]{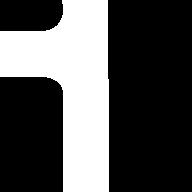}}
  \quad
  {\includegraphics[scale=0.4]{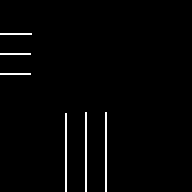}}
  \quad
  {\includegraphics[scale=0.4]{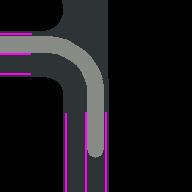}}
  \label{fig:bev}
  }
  \caption{a) Images from three frontal cameras located at the left, central, and right part of the vehicle, respectively. They were taken after the first few interactions of the agent in the CARLA simulation environment. 
  Each camera produces a 256x144 RGB image. 
  b) 
  \diff{The corresponding sparse trajectory visual input captured at the same frame.
  The points from the sparse trajectory and the highlighted vehicle position are plotted as circles with a radius of 10 pixels, using the same scale (pixels per meter) and perspective as the BEV representation. When the image is fed to the CGAN, it is represented with only one channel and a size of 192x192 pixels.} 
  c)
  The three channels of BEV image that our agent employs, computed at the same instant shown in (a). From left to right, the channels correspond to: desired route, drivable area, and lane boundaries. The last image shows all three channels combined in different colors.}
  \label{fig:cameras_all}
\end{figure*}

The BEV of a vehicle represents its position and movement in a top-down coordinate system 
\cite{chauffernet}. The vehicle's location, heading, and speed are represented by $p_t$, $\theta_t$, and $s_t$ respectively. 
The top-down view is defined so that the agent's starting position is always at a fixed point within an image (the center of it). 
Furthermore, it is represented by a set of images of size $W \times H$ pixels, at a ground sampling resolution of $\phi$ meters/pixel. 
The BEV of the environment moves as the vehicle moves, allowing the agent to see a fixed range of meters in front of it. For instance, the BEV representation for the vehicle whose three frontal cameras are shown in Fig.~\ref{fig:cameras} is given in Fig.~\ref{fig:bev}, where the desired route, drivable area and lane boundaries form a three-channel image.

\section{Agent}
\begin{figure*}[thpb]
  \centering
  \includegraphics[scale=0.38]{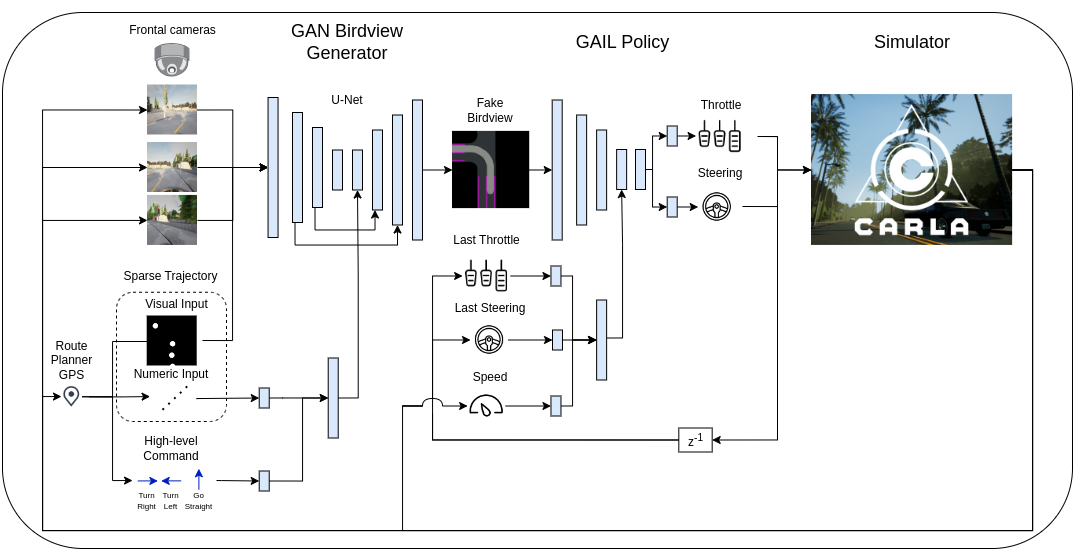}
  \caption{Hierarchical Generative Adversarial Imitation Learning (hGAIL) for policy learning with mid-level input representation. It basically consists of chained CGAN and GAIL networks, where the first one (CGAN) generates BEV representation from the vehicle's three frontal cameras, sparse trajectory and high-level command, while the latter (GAIL) outputs the acceleration and steering based on the predicted BEV input (generated by CGAN), the current speed and the last applied actions. Both CGAN and GAIL learn simultaneously while the agent interacts to the CARLA environment. The discriminator parts of both networks are not shown for the sake of simplicity.
  }
  \label{fig:hGAIL}
\end{figure*}

Our agent's architecture (Fig. ~\ref{fig:hGAIL}) is based on 
hierarchical Generative Adversarial Imitation Learning (hGAIL) for training policy and mid-level representation simultaneously. There are two main parts of hGAIL: the conditional GAN that generates the BEV representation based on input from the vehicle's frontal cameras, trajectory and high-level command; and the GAIL that learn the agent's policy by imitation learning based on input from the BEV representation generated by the first CGAN module, current vehicle's speed, and the last actuator values.

\subsection{BEV generation with CGANs}

The Conditional GAN module, used to transform the images from the frontal cameras into a top-down view representation, has two different networks named 
Discriminator and Generator, whose architectures can be seen in \diffa{Appendix \ref{apndx:gan_arch}}.

\subsubsection{Input representation}

The input for the CGANs corresponds to the 192x192 resolution RGB images from the three frontal \diff{cameras represented in Fig.~\ref{fig:cameras} and the sparse trajectory visual representation from Fig.~\ref{fig:traj_plot} that are stacked to generate a 10x192x192 image, i.e., with 10 channels.} The goal of the CGAN's generator is to translate this \diff{stack of images} into the 3-channel BEV representation seen in Fig.~\ref{fig:bev}.
In addition to this RGB input image, the discriminator also receives the 3x192x192 BEV image, which can come from either the generator as \textit{fake} or from the training set as \textit{real}.
Other inputs to the generator are the 
5 points from the sparse trajectory, one point behind the vehicle and 4 points ahead of it, \diff{represented as a vector} and the high-level command as a 4-dimensional one-hot encoding vector ("lane follow", "left", "right", "straight").

\diffa{The architectures of both Generator and Discriminator are described in Appendix \ref{apndx:gan_arch}.
}

\subsection{Policy learning with GAIL}
The generator in the GAIL module iteratively seeks the $\theta$ parameters of the policy $\pi_\theta(.|s)$ that minimizes
\eqref{eq:wgail00}, while the discriminator seeks to maximize it. To assist the agent's learning, loss terms for stimulating exploration are added as described after the representations for the input, output, and architecture are presented.

\subsubsection{Input representation}
The input $s$ to the agent's policy is a three-channels 192x192 image generated by the GAN network, corresponding to the mid-level BEV representation of the vehicle in its current position. In addition, the current vehicle's speed and the last value of the policy actuators (last acceleration and steering) are also fed as input further down in the network layers (to the first fully connected layer).

\subsubsection{Output representation}
The vehicle in CARLA has three actuators as: $
steering \in [-1,1], throttle \in [0,1]
$, and 
$brake \in [0,1]$.
Our agent's action space is $\mathbf{a} \in [-1,1]^2$, where the two components of $\mathbf{a}$ correspond to steering and acceleration. Braking occurs when acceleration is negative. In this way, by modeling brake and throttle with one dimension, the agent is not allowed to brake and accelerate simultaneously \cite{Petrazzini2021}.
Instead of using the Gaussian distribution for the policy's actions, common choice in model-free RL, we employ the Beta distribution $\mathcal{B}(\alpha, \beta)$ due to its bounded support, which allows us to model bounded continuous action distributions, usually found in real-world applications such as autonomous driving \cite{Petrazzini2021}, where the action space is not unbounded (i.e., the gas pedal can be actuated up to a certain limit). Besides, the policy loss $\mathcal{L}_P$ can be explicitly computed since clipping or squashing is not used to enforce input constraints (in the case of Gaussian distribution).
Furthermore, the Beta distribution allows the policy to act in extreme situations of vehicle driving, where sharp turns and sudden braking are necessary, as its parameters $\alpha$ and $\beta$, which are defined as outputs of the policy neural network $\pi_\theta$ and 
control the shape of the distribution, can be tuned to produce such characteristic vehicle behaviors.

\diffa{The architectures of both Generator and Discriminator are described in Appendix \ref{apndx:gail_arch}.}

\subsubsection{
\diffb{BC augmentation}
}

\diff{The generator loss in GAIL is also augmented with a Behavior Cloning (BC) loss $\mathcal{L}_{BC}$ 
($-\mathop{\mathbb{E}}_{\pi_E}[\log(\pi(a|s))]$) using the available expert samples. 
The underlying idea is to provide the training process of the policy with useful and informative gradients, especially when the discriminator is not yet fully trained. This concept and the practical implementation follows \cite{Jena2020}.}

\subsubsection{Encouraging Exploration}
During training, the agent is encouraged to explore the environment through two objectives, as in \cite{roach}:

\begin{equation}
 \mathcal{L}_{\mathrm{ent}} + \mathcal{L}_{\mathrm{exp}}
\label{eq:exploration}
\end{equation}
where: the first loss function corresponds to the entropy loss commonly used to promote exploration:
\begin{equation}
\mathcal{L}_{\mathrm{ent}} = -\lambda_{\mathrm{ent}} \cdot \mathrm{H}(\pi_\theta(.|s)),
\label{eq:ent_loss}
\end{equation}
Minimizing $\mathcal{L}_{\mathrm{ent}}$ means maximizing entropy and thus uncertainty for the policy distribution $\pi_\theta$, which stimulates the agent try more diverse actions since the policy distribution for a certain state $s$ does not become too certain too quickly in the process.
It also
drives the action (policy) distribution towards a uniform prior (which represents maximum entropy and uncertainty) since it is equivalent to minimizing the KL-divergence to the uniform distribution defined in the support of the Beta policy $[-1,1]$:
\begin{equation}
\mathrm{H}(\pi_\theta) = - \mathrm{KL}(\pi_\theta || \mathcal{U}(-1,1)),
\label{eq:ent_kl_loss}
\end{equation}

We can also bias the agent's learning with priors that signify meaningful behaviors for an autonomous vehicle and helps to improve and speed up the overall agent's training from scratch. 
This is accomplished with the following exploration loss $\mathcal{L}_{\mathrm{exp}}$ \cite{roach}:

\begin{equation}
\mathcal{L}_{\mathrm{exp}} = \lambda_{\mathrm{exp}} \cdot \mathbbm{1}_{\{T-N_{z}+1,...,T\}}(k) \cdot \mathrm{KL}(\pi_\theta(.|s) \left | \right | p_{z}),
\label{eq:exp_loss}
\end{equation}
where $\mathbbm{1}$ is the indicator function and $z \in \mathcal{Z}$ is the terminal event that finishes the episode. Some examples of events in $\mathcal{Z}$ would be collision, route deviation or the car being still or blocked for too long. 
$\mathcal{L}_{\mathrm{exp}}$ imposes a prior $p_z$ to the policy during the last $N_z$ steps of an episode ending with one of the events in $\mathcal{Z}$. The indicator function serves as a selection mechanism of the last steps in the episode. 
This $p_z$ promotes exploration as follows:
if $z$ is a collision, $p_z = \mathcal{B}(1,2.5)$ for the acceleration actuator, which encourages slowing down behavior; if the car is still, the acceleration prior is $p_z = \mathcal{B}(2.5,1)$, favoring increasing the vehicle's speed; if the vehicle deviates from the trajectory, a uniform prior $\mathcal{B}(1,1)$ is employed for the steering actuator \cite{roach}.

Thus, uniting the \diffb{Proximal Policy Optimization (PPO)} loss \diff{$\mathcal{L}_{P}$ \cite{ppoOriginal} that minimizes \eqref{eq:wgail00}, 
the Behavior Cloning loss $\mathcal{L}_{BC}$ \cite{Jena2020}}
\diffb{and the entropy and exploration terms in} \eqref{eq:exploration}, 
the total loss function for policy learning \diffb{of all GAIL-based agents in this work} is as follows:
\begin{equation}
\alpha \mathcal{L}_{BC} + (1 - \alpha) \mathcal{L}_{P}
+ \mathcal{L}_{\mathrm{ent}} + \mathcal{L}_{\mathrm{exp}},
\label{eq:bcgail_explo}
\end{equation}
\diffb{where $\alpha$ controls the weight of the BC term $\mathcal{L}_{BC}$ with respect to the PPO loss $\mathcal{L}_{P}$ during training. 
}

\section{Experimental Results}
\label{sec:result}
The goal of the vehicle is \diffa{to learn} to navigate autonomously in the city shown in Fig. \ref{fig:town01} using the hGAIL agent's architecture with mid-level BEV input generation, 
\diffa{and subsequently generalize its driving skill to a new town.}

\subsection{Collected data}
\label{sec:data}

The environment and trajectories are obtained from the CARLA Leaderboard evaluation platform \footnote{CARLA Autonomous Driving Leaderboard available at: https://leaderboard.carla.org/}. In particular, the \textit{town01} environment from this platform along with ten predefined trajectories are employed to generate the expert training set.

The expert dataset is constructed using a deterministic agent that navigates using a dense point trajectory and a classic PID controller \cite{chen2019learning}. The dense point trajectory provides many points at a fine resolution, whereas a sparse point trajectory consists of considerably fewer points, providing only a general sense of direction to the agent. As a result, the dense point trajectory is utilized to generate training data by the expert, whereas the sparse point trajectory is employed by the agent for more general guidance.

\begin{figure}[tb]
  \centering
  {\includegraphics[scale=0.30]{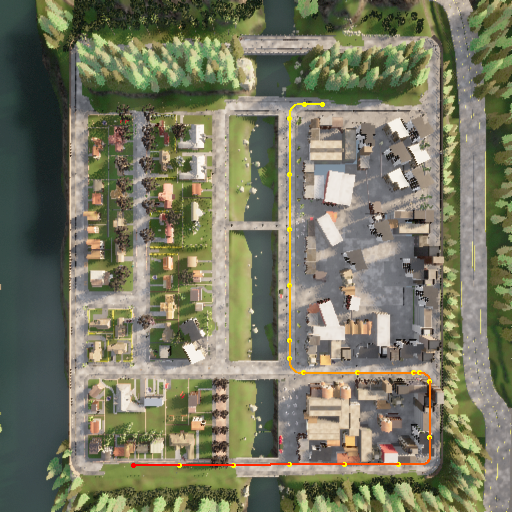}}
  \caption{\textit{Town01} environment of the agent, with one of the routes used to collect data by the expert. The highlighted path has 740 meters, 20 points in the sparse trajectory (shown as yellow dots) and 762 points in the dense point trajectory (not shown).} 
  \label{fig:town01}
\end{figure}

In Fig.~\ref{fig:town01}, one of the 10 routes executed by the expert to form the labeled training set of demonstrations is shown, where the line starting in yellow and ending in red represents the desired trajectory (not observable to the agent as it is). 
The sparse trajectory can be seen as yellow dots, generated every 50 meters traveled or when the vehicle is about to start a different movement (from \textit{straight} to \textit{turn} and vice-versa).

The ten trajectories of the training set were recorded at a rate of 10 hertz, resulting in 10 observation-action pairs per second.
For the shortest route of 1480 samples (average route of 2129 samples), it represents 2.5 minutes (3.5 minutes) of simulated driving. All the ten trajectories yielded a total of 21,287 training samples (30 GB of uncompressed data). The total set corresponds approximately to 36 minutes or 8km of driving.

\subsection{Training}
\label{exp:training}

The training was conducted using six parallel actors in a synchronous manner, with each actor running its own instance of the CARLA simulator. 
In the simulation, each episode begins with the vehicle at zero speed at a random starting point. The episode concludes upon the occurrence of any infraction, collision, or lane invasion, and a new episode begins with the vehicle located at a random point of the map to provide diversified experiences for each policy update.

At every $12,288$ environment interactions (steps), the agent's architecture is updated in a central computer: the parametrized policy using loss function \eqref{eq:bcgail_explo} is trained for 20 epochs using PPO (K=20), while the GAIL's Discriminator is trained for 2 epochs (J=2) on these $12,288$ samples; 
and the CGAN's Generator for BEV with the loss function \eqref{eq:cgan} is trained for 4 epochs, while its Discriminator for 4 epochs. This process corresponds to one training cycle of the full hGAIL. A new cycle will collect the next $12,288$ samples from all the actors, and execute the training as described above again.
As six parallel actors are used, 2,048 steps or environment interactions per actor are recorded, totalling 
the 12,288 environment interactions.
Thus, the episode does not have to end for a policy update to happen.
It is important to note that at any given moment, any of the six actors may be interacting with the environment in different parts of the environment.
Additional hyperparameters's values can be found in 
\diff{Appendix \ref{apndx:hyperparam}}.

It is worth noting that the \diff{GAN's generator} of hGAIL is \diff{pretrained on the fixed set of the ten expert trajectories (with 21,287 pairs of input and BEV targets) for 4 epochs in a supervised way.}
\diff{After this pretraining, the batch of the \textit{real} BEV samples for training the GAN's discriminator} evolves with the agent's training and corresponds exactly to the batch of $12,288$ samples collected by the six parallel actors. 
\diffa{These samples include the topdown view of the vehicle, computed by the simulator, which becomes the real BEV images which the GAN should learn to generate.}
This happens at every $12,288$ steps executed by all actors together and, thus, the training of the GAN \diff{for generating a} mid-level input representation \diff{is accomplished as the agent interacts with} the environment, similarly to a strategy for decoupling representation learning from reinforcement learning in \cite{rl_decoupling}.
The idea is that, once the GAN's training is turned off, the predicted BEV from the \diff{GAN's generator} could be used in \diffa{new city} where the \diffa{real} BEV is not available. \diff{The detailed algorithm for hGAIL training is shown in algorithm $(\ref{alg:hgail})$.}
\diff{
It is relevant to notice that
the training of the BEV representation will benefit from the online training samples collected as the agent interacts with the environment.
In this setting, many examples are generated where the agent \diffa{deviates} from the correct trajectory, automatically enriching the training of the BEV network.
}

\begin{algorithm}
\diff{
\small
\caption{Hierarchical Generative Adversarial Imitation Learning}\label{alg:hgail}
\textbf{Input:} Expert transitions buffer $\mathcal{B}_E$, $T$, $J$, $K$, $L$ \;
\textbf{Input parameters:} Actor $\theta$, Critic $\phi$, Discriminator $\omega$ \;
Pretrain CGAN BEV network $G(.)$ w/ samples from $\mathcal{B}_E$\;
\For{$episode = 1, 2, \dots$}{
    \For
    (\tcp*[f]{\small Collect env. samples})
    {$t = 1, 2, \dots, T$}{
        \tcp{\small $x_{t}$ are CGAN $G(.)$'s inputs; $y_{t}$ is the true BEV}    
        Choose action $a_t \sim \pi_{\theta} \left( G \left( x_t \right) \right)$;  $v_t \leftarrow V_{\phi} \left( G \left( x_t \right) \right)$\;
        $x_{t+1}, y_{t+1} \leftarrow act(a_t)$\; 
        Add $\left( x_t, y_t, G \left( x_t \right), a_t, v_t \right)$ to the buffer $\mathcal{B}_{\pi}$\;
    }
    \For(\tcp*[f]{\small Update GAIL discrim.}){$j = 1, 2, \dots, J$}
    {
        Sample $\{( y^{(i)}, a^{(i)})^{\pi}\}^{m}_{i=1}$ and $\{( y^{(i)}, a^{(i)})^{E}\}^{m}_{i=1}$ from policy transitions buffer $\mathcal{B}_{\pi}$ and expert transitions $\mathcal{B}_E$, respectively\;
        Update the policy discriminator parameters $w$ to increase $(\ref{eq:wgail00})$\;
    }
    Compute advantage $A_{t \in \{1,2, \dots, T\}}$ and add to policy transitions buffer $\mathcal{B}_{\pi}$\; 
    \For(\tcp*[f]{\small Update agent w/ PPO}){$k = 1, 2, \dots, K$}{
        Sample $\left \{\left(G \left( x^{(i)} \right), a^{(i)}, A^{(i)} \right)^{\pi}\right \}^{m}_{i=1}$ from 
        $\mathcal{B}_{\pi}$\;
        Update policy 
        parameters $\theta$ to minimize $(\ref{eq:bcgail_explo})$\;
    }
    Train CGAN BEV network w/ samples from $\mathcal{B}_{\pi}$ for $L$ epochs (both generator $G(.)$ and its discriminator)\;
}
}
\end{algorithm}

\subsection{Evaluation}
\diff{The main evaluation environment is \textit{town2}, shown in \diffa{Appendix \ref{apndx:town02}}. 
It was used to test how well the hGAIL agent can generalize its driving skills to unseen, new environments. All experiments below consider agents trained exclusively in \textit{town01} environment.}

\begin{figure}[tb]
  \centering
   \subfigure[\diff{With visual trajectory input}]{
    {\includegraphics[scale=0.45]{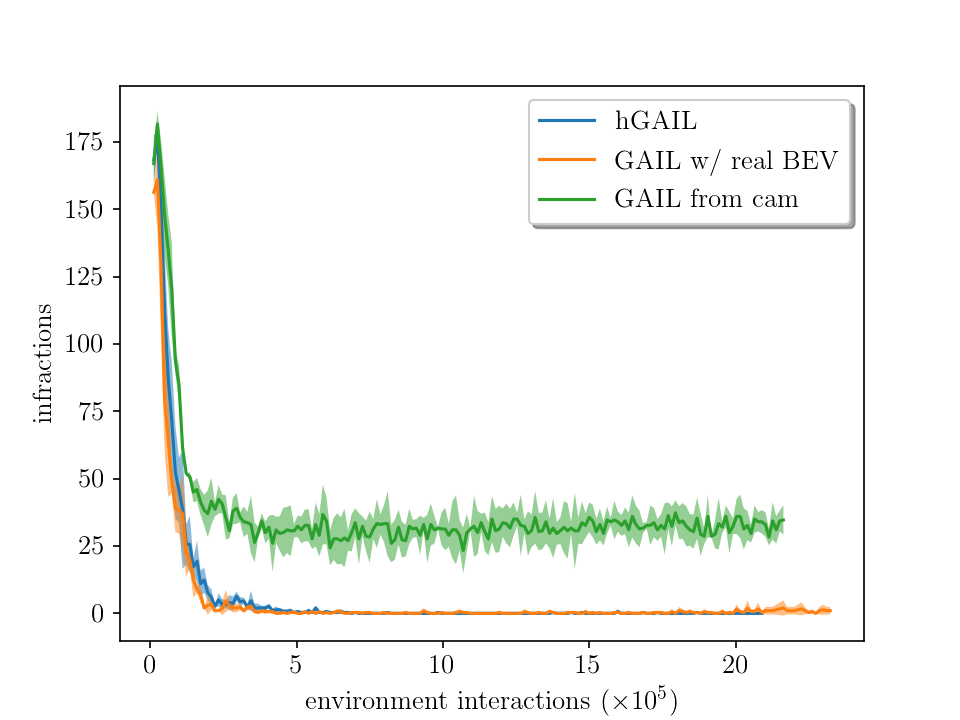}}
  }  
  \subfigure[\diff{Without visual trajectory input}]{
    {\includegraphics[scale=0.45]{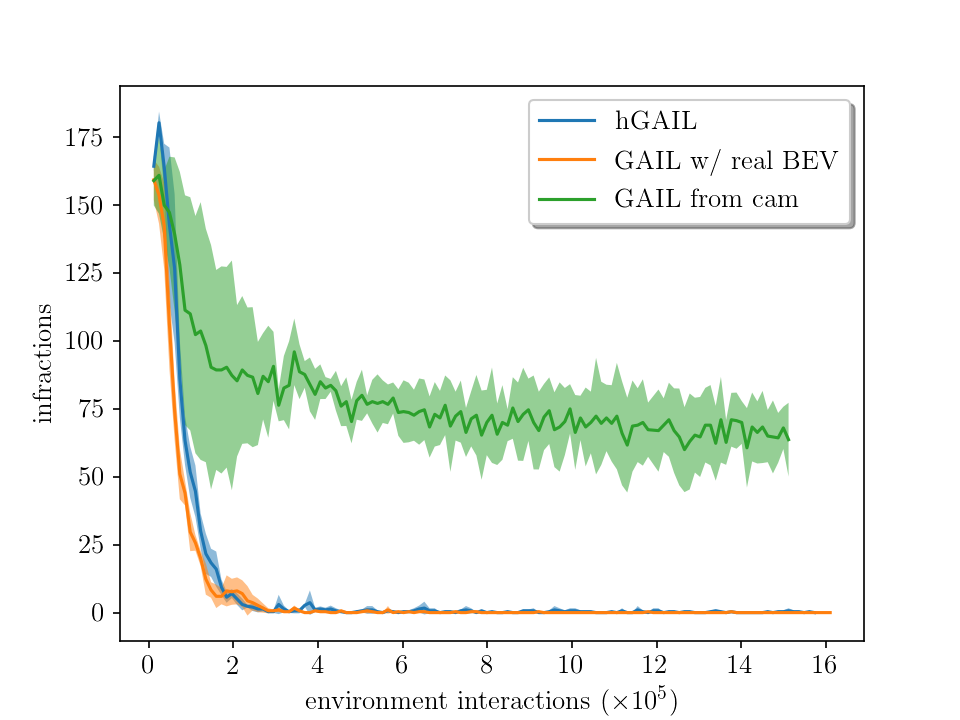}}
  }
  \caption{Number of committed infractions vs. environment interactions during training \diff{in \textit{town1} environment.} 
  \diff{The top (bottom) plot shows the results for the agents receiving (disregarding) the sparse trajectory as visual input (\textit{trajectory 1x192x192} in Fig.\ref{fig:gan}) in the corresponding CNN.}
  For each method (hGAIL, GAIL with real Bird's-Eye View, GAIL from cameras), the average performance of three runs is depicted considering a stochastic policy. The shaded area represents the standard deviation. 
  The GAIL from cameras agent fails to learn the task and keep the sum of committed infractions \diffa{close to} zero, while the \diffa{goal} of zero infractions is achieved by both hGAIL and GAIL with real BEV. 
  }
  \label{fig:infractionsGraph}
\end{figure}

The training progress can be seen in Fig.~\ref{fig:infractionsGraph} \diff{for \textit{hGAIL},
\textit{GAIL w/ real BEV}, and
\textit{GAIL from cameras}} 
agents. 
The second agent is trained directly \diff{on the real Bird's-Eye View image computed from the simulator,} while the last one is trained \diff{with input coming directly from the three frontal cameras and sparse trajectory, disregarding any Bird's-Eye View representation.}
The plot shows the average and standard deviation of the number of infractions for three runs for each \diff{agent's stochastic policy.}

\begin{figure}[tbp]
  \centering
  \subfigure[Evaluation]{
    {\includegraphics[scale=0.44]{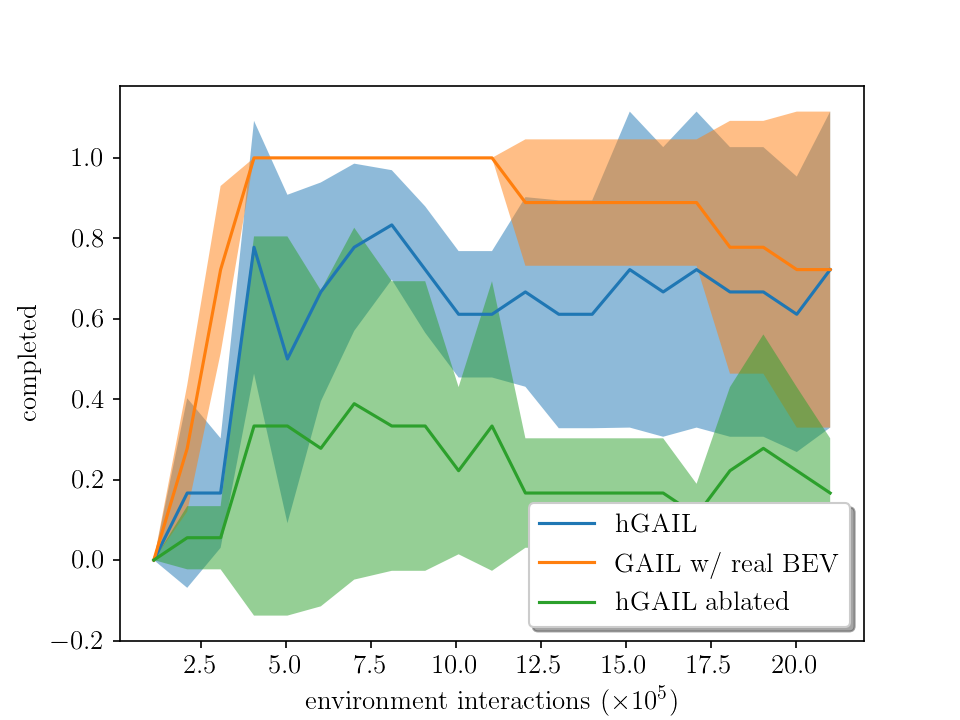}}
  }
  \caption{ \diff{Evaluation of agents in \textit{town2}, trained exclusively in \textit{town1}.
  The plot shows the
  percentage of completed routes from a total of six Leaderboard routes in \textit{town2} vs. environment interactions, averaged over three different runs, where each run entails a different agent trained only in \textit{town1}.
  For each method (hGAIL, GAIL with real Bird's-Eye View, hGAIL ablated), the average performance of three runs is depicted considering a deterministic policy. 
  The shaded area represents the standard deviation. Not shown in the plot, Behavior Cloning (BC) and GAIL from cameras agents fail to learn the task and
  complete any route (staying at 0\% if shown in the plot).
  Both hGAIL and \textit{GAIL with real BEV} agents are able to generalize the learning in \textit{town1} to \textit{town2}. The latter agent does not have to learn BEV, as it has always access to the true BEV.
  The \textit{hGAIL ablated} agent receives no visual input of the sparse trajectory, but only its numeric vector.
  }}
  \label{fig:completed_routes}
\end{figure}

The resulting deterministic policies of each agent \diff{trained in \textit{town1}} for all three runs are also evaluated \diff{in \textit{town2}} as training evolves, as shown in 
\diff{Fig.~\ref{fig:completed_routes}.
It shows
the average percentage of completed routes from a total of six Leaderboard routes in \textit{town2} as learning proceeds.
Each run uses a different agent trained exclusively in \textit{town1}.
Both hGAIL and GAIL with real BEV are able to generalize the learning in \textit{town1} to \textit{town2}.
Besides, the ablation on \textit{hGAIL} agent of the visual input of the sparse trajectory negatively affect the evaluation performance. In this context, it is worth noticing that if we would have pretrained its GAN network for more epochs on the expert dataset, we would improve the performance of the \textit{hGAIL ablated} (results not shown). This means that the sparse trajectory given as extra visual stimuli makes the GAN train faster.}
  
\begin{figure*}[tb]
  \centering

  \subfigure[top-right]{
    {\includegraphics[scale=0.35]{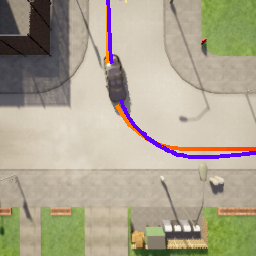}}
  }
  \subfigure[top-left]{
    {\includegraphics[scale=0.35]{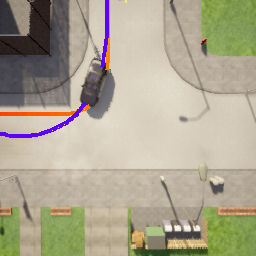}}
  }
  \subfigure[right-left]{
    {\includegraphics[scale=0.35]{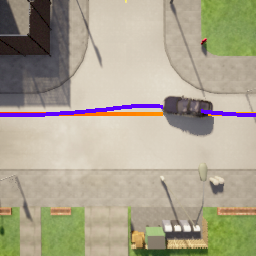}}
  }
  \subfigure[right-top]{
    {\includegraphics[scale=0.35]{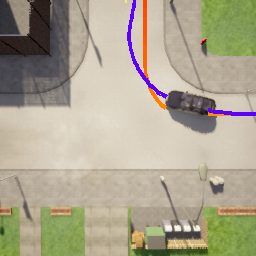}}
  }
  \caption{
    Agent's trajectories \diff{in \textit{town2} in blue color} generated by the deterministic policy after training \diff{in \textit{town1} (at epoch 100)} superimposed on the expert trajectory in orange color. At the same T intersection, 6 possible movements are possible: from top to right, top to left, right to left, right to top, 
    left to right (ommited) and left to top (ommited).
  } 
  \label{fig:trajectories_hgail}
\end{figure*}

\diff{In addition to the above three agents, other two agents were tested: the Behavior Cloning (BC) and GAIL from cameras agents.
The BC agent consists of basically substituting the GAIL policy for a BC policy, which receives the BEV prediction from a GAN. Both BC and its associated GAN were trained on the same set of ten demonstration trajectories in an offline manner.
Both of these agents are not shown in the plot, since they fail to learn to complete any route (staying at 0\% if shown in Fig.~\ref{fig:completed_routes}).}
After training, the agent was \diff{also} evaluated at a given T intersection and compared to the target given by the expert. Fig.~\ref{fig:trajectories_hgail} shows the resulting trajectories, with blue and orange denoting the agent's and expert's trajectories, respectively. It is worth noting that the \diff{policy's network in the hGAIL} agent receives as input only the generated (fake) BEV mid-level image, the current speed, and last applied actions for throttle and steering.
For instance, this BEV image corresponds to the topdown image with three channels from Fig.~\ref{fig:bev}. It is important to observe that the only information denoting the desired 
movement for the agent comes from the yellow desired route in the drivable red area. This yellow route occupies the whole lane in the BEV image, which leaves open how the agent will learn to turn at certain intersections. In other words, the agent's policy can not see directly the points in the sparse trajectory, as these points are fed to the GAN part of the architecture and not to the policy. 
This means that how we terminate the episode, \diff{such as through} infractions and lane invasion, will influence to a great extent the type of behavior the agent learns. Such an example can be seen in the turns of Fig.~\ref{fig:trajectories_hgail}, where the agent's trajectory does not match exactly with the expert's one. 
\diff{We conjecture that the behavior of the agent could have been made more similar to the expert if:
the network of the agent's policy would have received all the information that the expert receives as input (e.g., dense trajectory); the infractions would have been more strict; additional reward terms would have been added in the reward function to punish certain behaviors.}

The \diff{agent trained only on \textit{town1}} was also evaluated at every T intersection in \diff{\textit{town2}} environment, i.e., \diff{8} different T intersections, and compared to Behavior Cloning and GAIL from cameras agents. 
The latter corresponds to a GAIL agent with input directly from the vehicle's three frontal \diff{cameras instead of} the mid-level BEV input.
The results are summarized in Table~\ref{tab:intersect_bench}, whose lines presents the results for each possible turn out of 6 in total at a given T intersection (as shown in Fig.~\ref{fig:trajectories_hgail}). 
Thus, each turn,\diff{covering around 100 meters,} was evaluated in \diff{8} different T intersections, totalling \diff{48} experiments for each agent. 
The success percentage for each turn type is given in this table, where we can see that hGAIL can turn without failing in all intersections and for all turn types \diff{except for one \textit{top-right} intersection}, 
while BC fails \diff{50\%} 
of the times, 
and GAIL from cameras fails to learn most of the required driving behavior, succeeding only in \diff{4} 
turns out of \diff{48}. 
This ablation of the GAN from hGAIL (which is the GAIL from cameras) shows the need for learning the mid-level input representation to succeed in this complex task.
\diff{Additionally, notice that only hGAIL can finish complete trajectories (of around 1 km, as shown in Fig.~\ref{fig:completed_routes}), as both BC and GAIL from cameras fail in at least one turn type.}

 \begin{table}[]
    \centering
        \caption{Evaluation performance for 8 T Intersections and 6 type of turns in Town2}    
        \label{tab:intersect_bench}
    \begin{tabular}{lcccc}
    \hline 
    Turn type & BC & hGAIL & GAIL from cam. & hGAIL ablated
    \\ \hline
    Top-right & $0\% $ & $88\% $ & $0\% $ & $100\% $
    \\    
    Top-left & $75\% $ & $100\% $ & $0\% $ & $75\% $
    \\
    Right-left & $50\% $ & $100\% $ & $25\% $ & $100\% $
    \\
    Right-top & $88\% $ & $100\% $ & $13\% $ & $63\% $
    \\
    Left-right & $88\% $ & $100\% $ & $13\% $ & $100\% $
    \\
    Left-top & $0\% $ & $100\% $ & $0\% $ & $88\% $
    \\
    \hline
    All types & $50\% (24) $ & $98\% (47) $ & $8\% (4) $ & $88\% (42) $
    \\
    \hline
    \end{tabular}
    
\end{table}

\subsection{Mid-level representation learning}

\begin{figure*}[tb]
  \centering
  \subfigure[left]{
    {\includegraphics[scale=0.17]{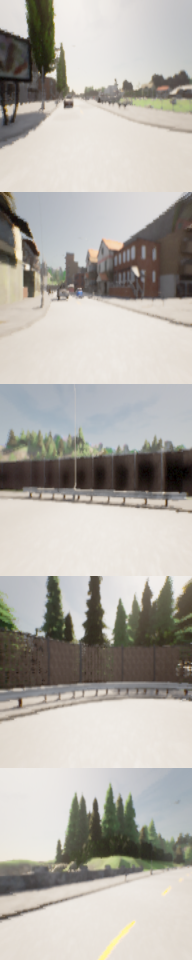}}
  }
  \subfigure[cent.]{
    {\includegraphics[scale=0.17]{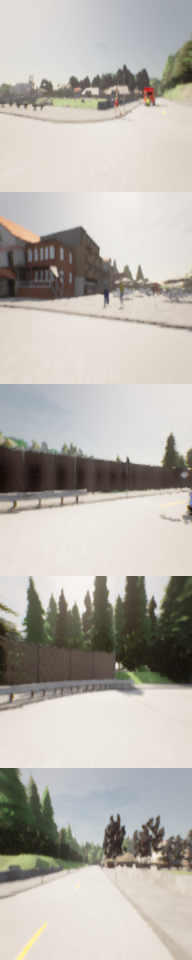}}
  }
  \subfigure[right]{
    {\includegraphics[scale=0.17]{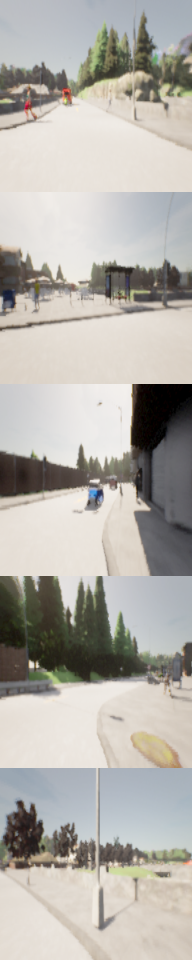}}
  }
  \subfigure[traj.]{
    {\includegraphics[scale=0.17]{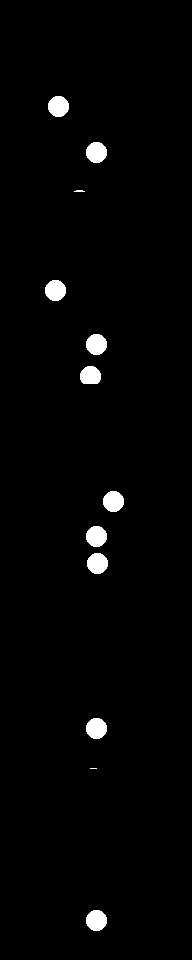}}
  }
  \subfigure[target]{
    {\includegraphics[scale=0.17]{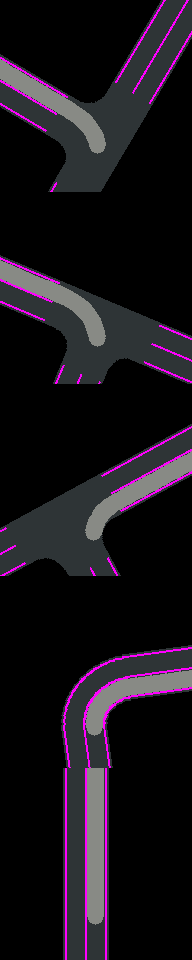}}
  }
  \subfigure[1 cy.]{
    {\includegraphics[scale=0.17]{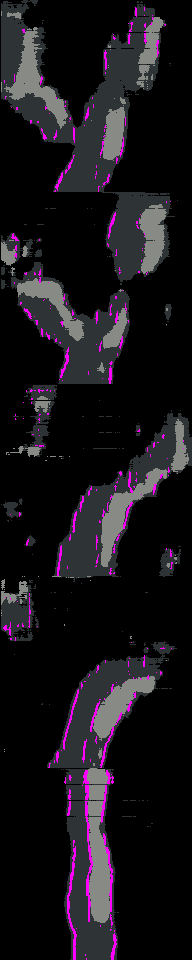}}
  }
  \subfigure[10 cy.]{
    {\includegraphics[scale=0.17]{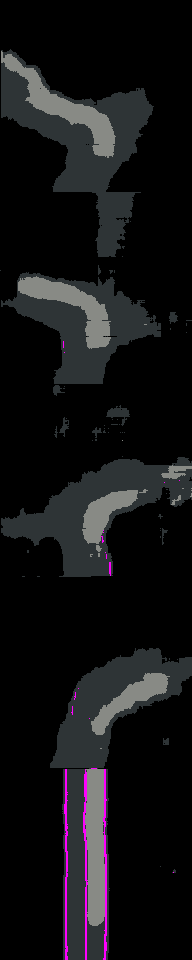}}
  }
  \subfigure[20 cy.]{
    {\includegraphics[scale=0.17]{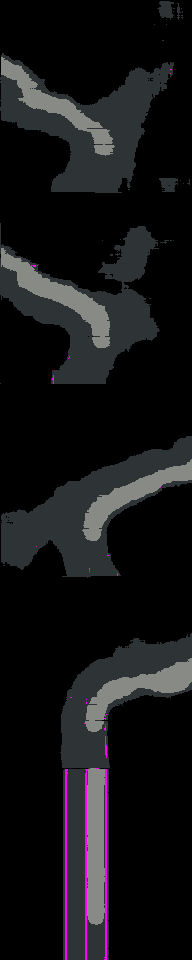}}
  }
  \subfigure[50 cy.]{
    {\includegraphics[scale=0.17]{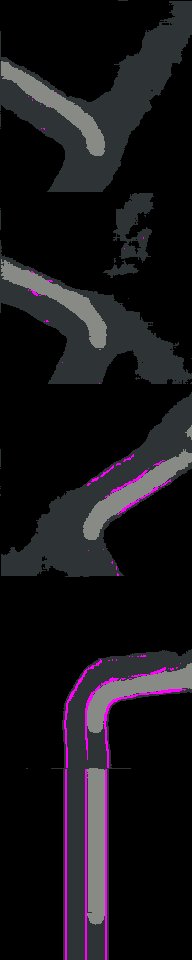}}
  }
  \subfigure[90 cy.]{
    {\includegraphics[scale=0.17]{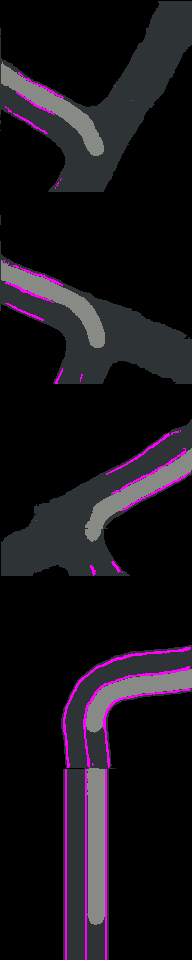}}
  }
  \caption{BEV generation \diffa{in \textit{town2}} as the agent goes through training in \diffa{\textit{town1}}.
  \diff{The first three columns shows the images from the cameras attached to the front of the vehicle.}
  \diffa{The fourth column shows the sparse trajectory from the route planner.}
  \diff{The fifth} column shows five BEV images computed by the simulator and are considered the target output. 
  The following columns show the BEV images generated by the GAN from the agent's architecture as it undergoes training, at: 12,288 environment steps (1 cycle), \diff{122,880 environment steps (10 cycles)}, \diff{245,760 environment steps (20 cycles)}, \diff{614,400} environment steps (\diff{50} cycles), and \diff{1,105,920 environment steps (90 cycles)}. One cycle is similar to the concept of epoch, and consists of the full training of hGAIL using the last 12,288 steps collected; 
  \diffa{see Section~\ref{exp:training} for more details.}
  }
  \label{fig:gan_eval}
\end{figure*}

Here, we \diff{present} some results of the Bird's-Eye-View representation \diff{learned} by the GAN's generator \diff{of} the hGAIL agent's architecture.
\diff{While this GAN learns in \textit{town01}}, 
Fig.~\ref{fig:gan_eval} shows
the evolution \diff{of} the representations of five different \diff{vehicle's positions taken} in \diff{\textit{town02}} at different training \diff{epochs of the agent in \textit{town1}}, where each row corresponds to a different particular position of the vehicle in the \diff{\textit{town02}} environment. 
\diff{The first four columns are the input to the GAN's generator, consisting of the images from the three frontal cameras and the sparse trajectory given as an image.}
\diff{The fifth} column corresponds to the targets \diff{(labels)}, i.e., the BEV generated by the simulator, which is used to train the GAN's generator. The other columns show the mid-level \diff{BEV} representation evolving from a poor prediction at cycle 1 (after 12,288 environment steps) to a good enough prediction at \diff{cycle 32. It is worth noticing that the GAN has never seen \textit{town02}, and was trained only on \textit{town01}.}


\subsection{Dynamic environment}

\diffa{
Here, we extend the hGAIL agent's architecture to allow for environments with traffic lights, pedestrians and other vehicles.
Initially, to make the agent able to learn to respect traffic lights, it was necessary to 
employ importance weighting in the BC term ($\mathcal{L}_{BC}$) of the loss function in (\ref{eq:bcgail_explo}) so that each sample in the expert set is weighted according to its importance. 
These weights were found by computing the inverse of the output of a Kernel Density Estimator (KDE) applied on the target outputs (action values), so that less frequent actions (e.g., accelerating as the light turns green) are given more weight in the BC loss. 
\diffb{The KDE used a Gaussian Kernel with bandwidth equal to 0.2.}
Additionally, three extra binary inputs were used for the GAIL policy in hGAIL, denoting the detection of traffic lights, pedestrians and vehicles in front of the cameras.
\diffb{Also, the weight of the BC term in \eqref{eq:bcgail_explo} was set to $\alpha=0.012$,
and the number of discriminator epochs ($J$) to 4.
}
}

\begin{figure}[tb]
  \centering
  \includegraphics[scale=0.5]{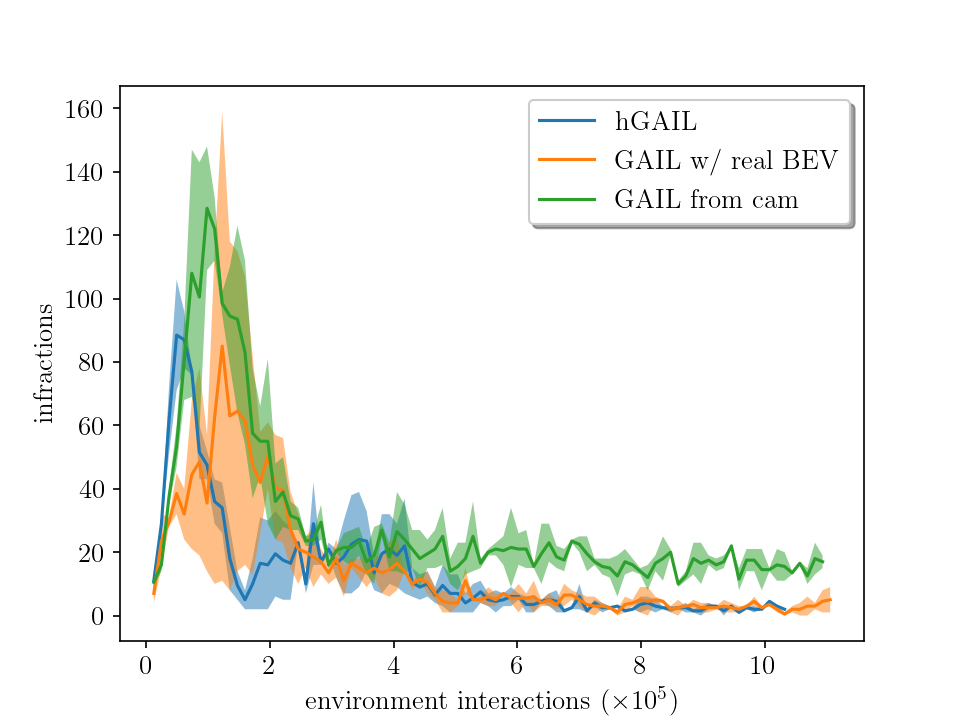}
    \caption{
    \diffa{Number of infractions for hGAIL, GAIL from real BEV, and GAIL from camera agents during training in \textit{town1} with pedestrians, vehicles and traffic lights.}
  }
    \label{fig:train_dynamicenv}
\end{figure}

\diffa{
The expert dataset was built similarly to the previous experiment, with the same trajectories, as described in Section \ref{sec:data}, except that now the expert respects traffic lights and breaks when pedestrians and other vehicles appear close enough ahead of its vehicle, which increases the dataset size.
We trained two agents from scratch for each method, i.e.,  hGAIL, GAIL with real BEV, \diffb{and GAIL from camera.
}
The last two methods also employed importance weighting and three extra inputs as done in hGAIL, but processed inputs directly from the real BEV of the scene 
\diffb{for the second method, and from the three frontal cameras for the last method.}
The training evolution can be seen in Fig.~\ref{fig:train_dynamicenv}, where the first two agents converge to a low infraction rate in \textit{town01}. 
The evaluation results can be seen in Table~\ref{tab:dynamic} for both training \textit{town1} and test \textit{town2} environments, showing the mean and the standard deviations 
\diffb{over all routes in each town and the two agents trained for each method}.
}
\diffa{Even though the GAIL agent has a slightly better \textit{Route Completion}, its \textit{Driving Score} (22\%) is worse than the score (45\%) of the hGAIL agent in \textit{town02} because the former commits more infractions and collisions.
It could be that the noisy predicted BEV that hGAIL uses as input to the policy makes it more robust when learning to avoid infractions in comparison to the GAIL with real BEV agent.
We can also notice that the expert fails sometimes. Particularly, hGAIL surpass the expert for the \textit{Red Light Infraction} in \textit{town01}, and it gets similar performance to the expert in other metrics.}
\diffb{The GAIL agent trained with input directly from camera presented the worst performance, with 28\% and 5\% average driving scores for \textit{town01} and \textit{town02}, respectively.}

\begin{table*}[h!]
\begin{threeparttable}
\centering
\caption{
\diffa{Driving performance and infraction analysis for different agents on Town01 and Town02.}
}
\begin{tabular}{|c|c|c|c|c|c|c|c|c|}
\hline
\multirow{2}{*}{\textbf{Agent}} & \multirow{2}{*}{\textbf{Town}} & \multicolumn{7}{c|}{\textbf{Metrics}}
\\
\cline{3-9}
& & \textbf{Driving} & \textbf{Route} & \textbf{Infraction} & \textbf{Collisions} & \textbf{Collisions} & \textbf{Red Light} & \textbf{Agent} \\
& & \textbf{Score} & \textbf{Completion} & \textbf{Penalty} & \textbf{Vehicle} & \textbf{Pedestrian} & \textbf{Infraction} & \textbf{Blocked} \\
& & \textbf{(\%, $\uparrow$)\tnote{a}} & \textbf{(\%, $\uparrow$)} & \textbf{(\%, $\uparrow$)} & \textbf{(\#/Km, $\downarrow$)} & \textbf{(\#/Km, $\downarrow$)} & \textbf{(\#/Km, $\downarrow$)} & \textbf{(\#/Km, $\downarrow$)} \\
\hline
\multirow{2}{*}{\textbf{Expert}} & \textit{town01} & 86 $\pm$ 9 & 98 $\pm$ 2 & 87 $\pm$ 8 & 0.0 $\pm$ 0.0 & 0.0 $\pm$ 0.0 & 0.62 $\pm$ 0.43 & 0.21 $\pm$ 0.21 \\
& \textit{town02} & 46 $\pm$ 16 & 82 $\pm$ 1 & 60 $\pm$ 18 & 0.31 $\pm$ 0.14 & 0.0 $\pm$ 0.0 & 2.03 $\pm$ 1.32 & 1.17 $\pm$ 0.29 \\
\hline
\multirow{2}{*}{\textbf{hGAIL}} & \textit{town01} & 82 $\pm$ 1 & 95 $\pm$ 6 & 86 $\pm$ 4 & 0.14 $\pm$ 0.14 & 0.09 $\pm$ 0.09 & 0.48 $\pm$ 0.11 & 0.26 $\pm$ 0.26 \\
& \textit{town02} & 45 $\pm$ 2 & 66 $\pm$ 8 & 65 $\pm$ 3 & 0.98 $\pm$ 0.49 & 0.17 $\pm$ 0.17 & 2.24 $\pm$ 0.95 & 2.00 $\pm$ 0.51 \\
\hline
\multirow{2}{*}{\textbf{GAIL}} & \textit{town01} & 67 $\pm$ 1 & 97 $\pm$ 0 & 70 $\pm$ 1 & 0.76 $\pm$ 0.27 & 0.05 $\pm$ 0.05 & 0.53 $\pm$ 0.09 & 0.13 $\pm$ 0.0 \\
& \textit{town02} & 22 $\pm$ 2 & 72 $\pm$ 1 & 31 $\pm$ 5 & 3.93 $\pm$ 1.13 & 0.08 $\pm$ 0.08 & 2.24 $\pm$ 0.93 & 1.62 $\pm$ 0.13 \\
\hline
\textbf{\diffb{GAIL}} & \textit{town01} & 28 $\pm$ 1 & 50 $\pm$ 6 & 61 $\pm$ 10 & 0.36 $\pm$ 0.0 & 0.35 $\pm$ 0.35 & 2.63 $\pm$ 0.46 & 8.32 $\pm$ 4.02 \\
\textbf{from cam}& \textit{town02} & 5 $\pm$ 2 & 10 $\pm$ 3 & 48 $\pm$ 4 & 10.76 $\pm$ 6.80 & 0.62 $\pm$ 0.62 & 0.62 $\pm$ 0.62 & 18.00 $\pm$ 3.75 \\
\hline
\end{tabular}
\begin{tablenotes}
\item[a]\diffa{ The \textit{Driving Score}, a metric from CARLA Leaderboard, is calculated by the product of \textit{Route Completion} (the percentage of the route distance completed) and an \textit{Infraction Penalty} which accounts for all violations committed. For instance, if an agent ran a red light and collided with a pedestrian during a route and the penalty for running one red light was 0.7 and for colliding with a pedestrian is 0.5, the penalty would be $0.7 \times 0.5 = 0.35$.
}
\end{tablenotes}
\end{threeparttable}
\label{tab:dynamic}
\end{table*}

\begin{figure}[tb]
  \centering
  \includegraphics[scale=0.5]{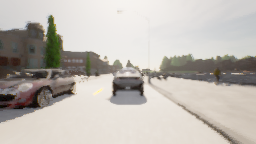}
  \includegraphics[scale=0.5]{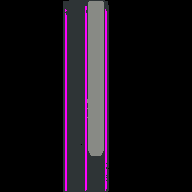}
    \caption{
    \diffa{BEV generation in a scene with two cars captured by the vehicle's camera. Left: central camera image input to hGAIL. Right: predicted BEV output by the CGAN in hGAIL.
    }
  }
  \label{fig:bev_dynamic}
\end{figure}

\diffa{In Fig.~\ref{fig:bev_dynamic}, we can observe that the hGAIL agent can generate the three BEV channels for navigation in spite of the pedestrians and vehicles that may pass in front of the cameras. Thus, it learns to predict what is the road ahead in the BEV representation as if the vehicle or pedestrian would not be in front of the camera blocking the passage and the view.
}

\section{Conclusion}
\label{sec:conclusion}

In this work, the hGAIL architecture was proposed to solve the autonomous navigation of a vehicle \diff{in the CARLA simulation environment} in an end-to-end approach, connecting sensory perceptions to low-level actions directly with neural networks (sensory-motor coupling), while learning mid-level input representations of the agent's environment.
hGAIL is an hierarchical Adversarial Imitation Learning architecture composed of two main modules: the CGAN which generates the Bird's-Eye View (BEV) representation from the three frontal cameras of the vehicle, desired trajectory and high-level command, which is a mid-level (more abstract) input representation of the scene in front of the vehicle; and the GAIL which learns to control the vehicle based mainly on the BEV predictions from the CGAN as input.

Both GAIL and CGAN in hGAIL learns simultaneously 
to control the agent and generate the input representations, respectively. 
The learning takes place 
in an urban city
\diffa{initially}
without pedestrians or other cars, but with dynamic routes that can change the path on the fly. 
\diffa{In the following experiment, hGAIL is extended so that the agent can learn to respect traffic lights and deal with the detection of pedestrians and vehicles. 
}%
Our experiments have shown that the mid-level input generated by CGAN is essential for the learning task as the GAIL exclusively from cameras (without BEV) fails to even learn the task, keeping a high-infraction rate through training. The BC agent with its associated GAN can complete some turns in the trajectories, but not consistently as hGAIL can. In fact, hGAIL, after \diff{training exclusively on ten expert trajectories from one city, was able to generalize its navigation skills to a new, unseen city, and complete 98\% of all 48 possible} turns in the \diff{eight} T intersections from \diff{that new} city.

This work has demonstrated the usefulness of mid-level BEV input for realistic navigation scenarios, but also that this input representation can be learned concomitantly with the agent's policy training. Thus, the BEV generation is learned with the same data distribution used to train the agent's policy, \diff{i.e., the expert dataset and the data generated by the agent}. In future work, we are interested in extending the method
making the policy generalize
\diffa{to other weather conditions and}
\diff{other new cities; and sim2real experiments}. 
\diffa{Also, we would like to investigate more closely the rarer situations that lead to infractions, proposing a more refined architecture to handle these difficult cases, as well as improve the quality of the expert.}

\addtolength{\textheight}{-0cm}
\section*{APPENDIX}

\subsection{Generative Adversarial Imitation Learning (GAIL)}
\label{apndx:gail}

In 
GAIL \cite{Ho2016}, basically, there are two components that are trained iteratively in a min-max game: a discriminative classifier $D$ is trained to distinguish between samples generated by the learning policy $\pi$ and samples generated by the expert policy $\pi_E$ (i.e., the labelled training set); and the learning policy $\pi$ is optimized to imitate the expert policy $\pi_E$. Thus, in this game, both $D$ and $\pi$ have opposite interests: $D$ feeds on state-action pair $(s,a)$ and its output seeks to detect whether $(s,a)$ comes from learning policy $\pi$ or expert policy $\pi_E$; and $\pi$ maps state $s$ to a probability distribution over actions $a$, learning this mapping by relying on $D$'s judgements on state-action samples (i.e., $D$ informs how close $\pi$ is from $\pi_E$).
Mathematically, GAIL finds a saddle point $(\pi,D)$ of the expression:
\begin{equation}
\mathop{\mathbb{E}}_{\pi_E}[D(s,a)] - \mathop{\mathbb{E}}_\pi[D(s,a)] - \lambda H(\pi) - \lambda_2 L_{gp}, 
\label{eq:wgail2}
\end{equation}
where $D: S \times A \rightarrow (0,1) $, $S$ is the state space, $A$ is the action space; $\pi_E$ is the expert policy; $H(\pi) $ is a policy regularizer controlled by $\lambda >= 0$ \cite{Bloem2014}.
\diff{The discriminator will try to increase (\ref{eq:wgail2}), while $\pi$ seeks to minimize it; and $L_{gp}$ is a loss that penalizes the gradient's norm,
constraining the discriminator network to the 1-Lipschitz function space, according to \cite{gulrajani2017improved}.}
The above equation is the Wasserstein loss, used
to alleviate vanishing gradient and mode collapse problems, using
the Wasserstein distance \cite{gulrajani2017improved} between the policy 
distribution and expert distribution, as also done in \cite{Zhang_2020,li2017infogail}.
It measures the minimum effort to move one distribution to the place of the other, yielding a better feedback signal than the Jensen-Shannon divergence.
Both $D$ and $\pi$ can be represented by deep neural networks. In practice, a training iteration for $D$ uses Adam gradient-based optimization \cite{Kingma2014} to increase (\ref{eq:wgail2}), and in the next iteration, $\pi$ is trained with any on-policy gradient method such as Proximal Policy Optimization (PPO) \cite{ppoOriginal}
to decrease (\ref{eq:wgail2}).

\subsection{CGAN Network Architecture}
\label{apndx:gan_arch}
\begin{figure}[thpb]
  \centering
  \includegraphics[scale=0.39]{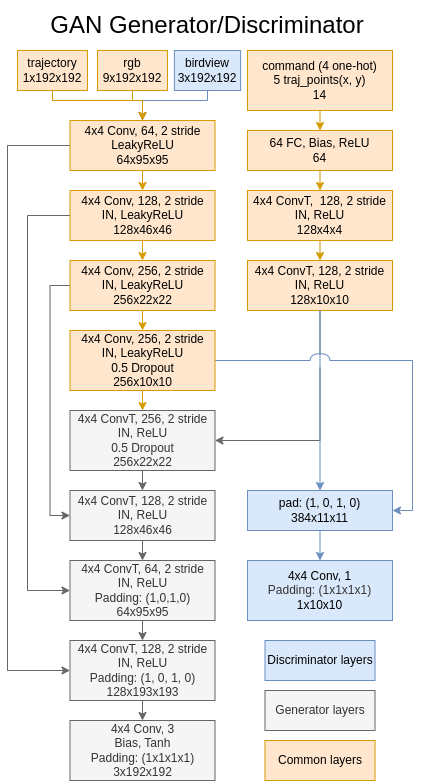}
  \caption{Conditional GAN architecture for generating the BEV input representation. The Generator and the Discriminator are separate networks which do not share parameters: the figure was made to not repeat equivalent layers when describing both networks. 
  The generator corresponds to the U-net at the left side of Fig.~\ref{fig:hGAIL} and aims at translating RGB \diff{10x192x192} images from the vehicle's frontal cameras \diff{and sparse trajectory} to BEV mid-level input representation (3x192x192 images). 
  }
  \label{fig:gan}
\end{figure}

\diffa{Both CGAN networks' architectures are} presented in Fig.~\ref{fig:gan}, where the common layers in orange refer to layers existing in both the Generator and the Discriminator. Notice that they are separate networks which do not share parameters: the figure was made to not repeat equivalent layers when describing both networks. 
\subsubsection*{Generator}
It can be seen in this figure and also in Fig.~\ref{fig:hGAIL} that the CGAN's generator is a U-Net \cite{unet}, usually employed for image translation or segmentation.
Further, while the image is processed by convolution layers, the other perceptual inputs (trajectory and command, second column in the figure) are processed by two fully connected layers followed by two transposed convolution layers which upsample their input to reach the desired resolution so that it can be merged with the last orange 256x10x10 layer in the left column.
The next transposed convolution grey layer (256x22x22) merges information coming from the frontal cameras's RGB images (left column) and the trajectory points plus the command (right command) for the generator network.
Its final output is 3x192x192, corresponding to the three-channels BEV translated image.
\subsubsection*{Discriminator}
The discriminator is also conditioned on the RGB images from the frontal cameras \diff{and the trajectory visual representation}, which \diff{are} merged to the (fake/real) BEV image, totalling \diff{13x192x192} input to the first convolutional layer of the discriminator. The other perceptual inputs (right column) are processed similarly to the generator until it merges in a new 384x11x11 layer (in blue) with information coming from the images (256x10x10, left column). The final output corresponds to the one given by PatchGAN.

\subsection{GAIL policy architecture}
\label{apndx:gail_arch}
\begin{figure}[thpb]
  \centering
  \includegraphics[scale=0.39]{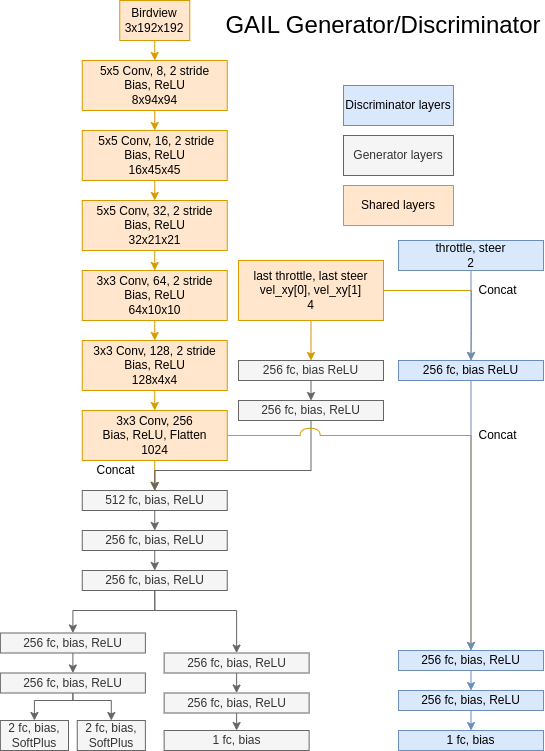}
  \caption{GAIL architecture for policy learning, corresponding to the Generator network at the right side of Fig.~\ref{fig:hGAIL} and the Discriminator responsible for producing the reward signal. The Generator, $\pi_\theta(a|s)$, receives the predicted BEV image from the GAN's generator, the last agent's actions (throttle, steer), and the current speed as input (which forms the observation $s$ of the policy), and outputs the $\alpha$ and $\beta$ parameters of the Beta distribution for both steering and action with the SoftPlus activation function. The Value function $V_\phi(s)$ shares the Generator network's layers until it branches into a separate head with more two hidden layers and a linear output unit. The Discriminator 
  $D_\omega(s,a)$ receives the actions \textit{throttle} and \textit{steer} in addition to the observation $s$ and has a linear output unit. Notice that features from the last convolutional layer are flattened before they are merged (\textit{concat}) with other information into FC (fully connected) layers.
  }
  \label{fig:gail}
\end{figure}

The agent's policy part, which corresponds to the right side of Fig.~\ref{fig:hGAIL},
has the architecture shown in Fig.~\ref{fig:gail}. The discriminator layers are also shown, even though both network's weights are not shared, as in the previously presented CGAN architecture. The only shared part corresponds to the layers between the Generator and the value function $V_\phi(.)$ until the main branch splits into two heads: one for the actions \textit{steering} and \textit{throttle} for the generator (with 2 \textit{softplus} units that outputs the $\alpha$ and $\beta$ parameters of a Beta distribution, for each action); and another for the value of state $s$, given by a linear unit. The discriminator $D(s,a)$ receives an action $a$ in addition to the observation $s$ and maps to a linear output unit, whose output value is employed as reward when training with PPO.


\subsection{Hyperparameters}
\label{apndx:hyperparam}

\diffa{Tables ~\ref{tab:hyperparamsGAIL} and
Table~\ref{tab:hyperparamsGAN} show the hyperparameter values for the GAIL and GAN parts of hGAIL, respectively}. 

 \begin{table}[h!]
    \centering
        \caption{Hyperparameters for GAIL}
        \label{tab:hyperparamsGAIL}
    \begin{tabular}{lc}
    \hline
    Description & Value
    \\ \hline
    Parallel environments ($N$) & $6$
    \\
    Initial adam step size (lr) & $2.0 \times 10^{-5}$
    \\
    Adam step size exponential decay ($\lambda_{lr}$) & $0.96$
    \\
    Number of PPO epochs (K) & $20$
    \\
    Mini-batch size (m) & $256$
    \\
    Discount ($\gamma$) & $0.99$
    \\
    GAE parameter ($\lambda$) & $0.9$
    \\
    Clipping parameter ($\epsilon$) & $0.2$
    \\
    Value Function clipping parameter ($\epsilon_{vf}$) & $0.2$
    \\
    Value Function coefficient ($c_{1}$) & $0.5 $
    \\
    Entropy coefficient ($\lambda_{\mathrm{ent}}$) & $0.01$
    \\
    Exploration coefficient ($\lambda_{\mathrm{exp}}$) & $0.05$
    \\
    Timesteps per epoch (T) & $12288$
    \\
    Weight of BC in GAIL loss ($\alpha$) & $0.004$
    \\
    Discriminator adam step size (lr) & $2.5 \times 10^{-4}$
    \\
    Number discriminator epochs ($J$) & $2$
    \\
    \hline
    \end{tabular}
\end{table}

 \begin{table}[h!]
    \centering
        \caption{Hyperparameters for CGAN }    
        \label{tab:hyperparamsGAN}
    \begin{tabular}{lc}
    \hline
    Description &  Value
    \\ \hline
    Adam step size (lr) & $2.0 \times 10^{-4}$
    \\
    Number of GAN epochs (L) & $4$
    \\
    Mini-batch size (m) & $32$
    \\
    Patch size ($\gamma$) & $(10, 10)$
    \\
    Resize ($\lambda$) & $(192, 192)$
    \\
    Lambda pixel ($\epsilon$) & $100$
    \\
    \hline
    \end{tabular}
\end{table}


\subsection{Training progress}
\label{apndx:training}

\begin{figure}[thpb]
  \centering
   \subfigure[]{
    {\includegraphics[scale=0.59]{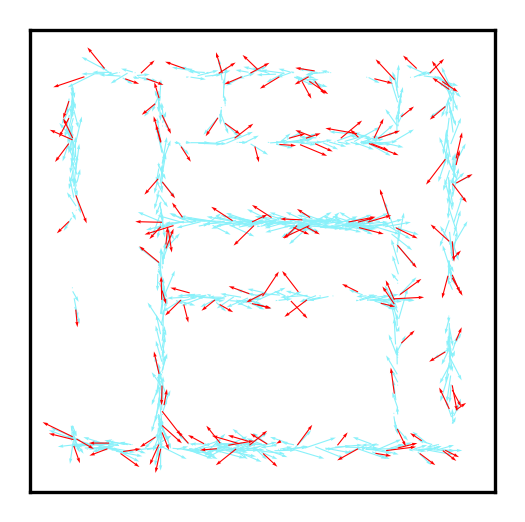}}
  }
  \subfigure[]{
    {\includegraphics[scale=0.59]{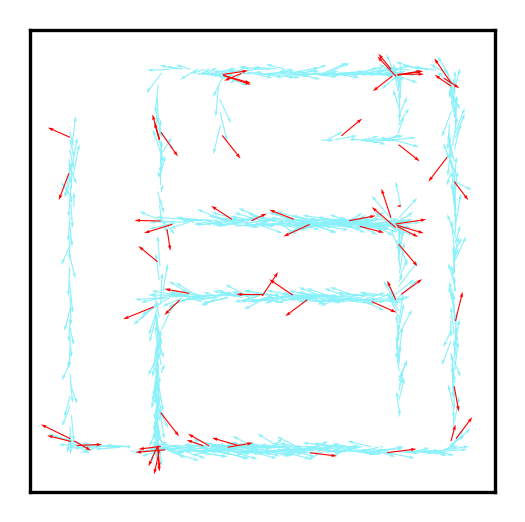}}
  }
  \subfigure[]{
    {\includegraphics[scale=0.59]{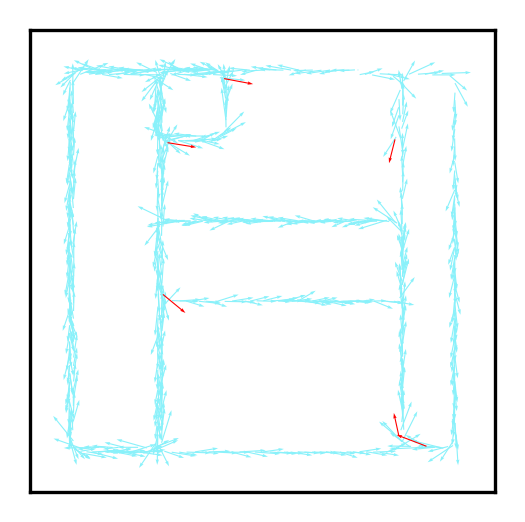}}
  }

  \caption{The vehicle's trajectory in \textit{town1}, in \diff{blue}, 
  during different moments of the training process. 
  In the early training iterations, errors, marked in red color, are common. As training proceeds, less and less mistakes happen\diff{: in (a), until 86,000 environment interactions; in (b) from 61,000 to 147,000 interactions; in (c) from 245,000 to 331,000 interactions}.
    }
    \label{fig:trainVectorFieldImages}
\end{figure}

The evolution of training for \diff{the hGAIL agent in \textit{town1}} can also be seen in Fig.~\ref{fig:trainVectorFieldImages}, where the whole trajectory throughout the city is plot at three different moments in training. Early in the training process, the infractions or errors, given by red triangles, are frequent. These infractions decrease as learning proceeds.


\subsection{Evaluation environment}
\label{apndx:town02}
\begin{figure}[thpb]
  \centering
  {\includegraphics[scale=0.35]{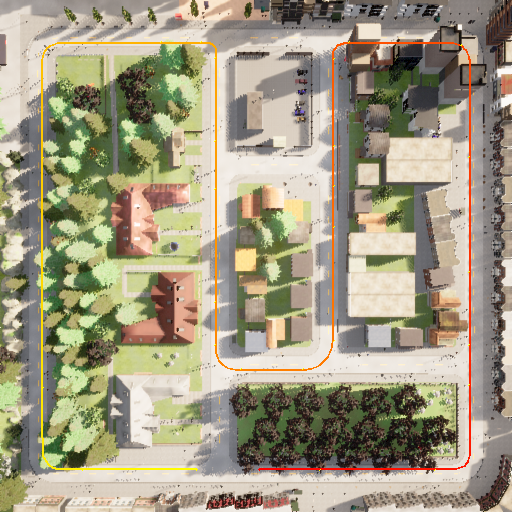}}
  \caption{\diff{\textit{Town02} \diffa{evaluation city}, with one of the routes used to test agents trained in \textit{town01}. The highlighted path has 1010 meters, 29 points in the sparse trajectory (shown as yellow dots) and 1030 points in the dense point trajectory (not shown).}} 
  \label{fig:town02}
\end{figure}

Town02 in Fig.~\ref{fig:town02} is the environment used for testing the generalization capabilities of the trained agents.

\section*{ACKNOWLEDGMENT}
The authors acknowledge CAPES (Coordination for the Improvement of Higher Education Personnel) - Brazil for financial support.

\bibliographystyle{IEEEtran}

\bibliography{IEEEabrv,conference}

\begin{IEEEbiography}[{\includegraphics[width=1in,height=1.25in,clip,keepaspectratio]{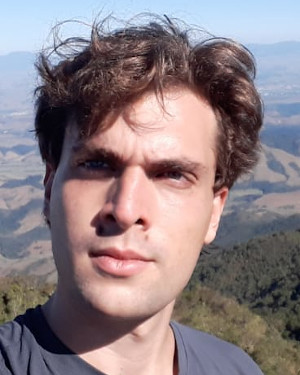}}]{Gustavo Claudio Karl Couto}
received the master’s degree in automation and systems engineering at Federal University of Santa Catarina in 2023 and graduated from the Militar Institute of Engineering in 2018 with a degree in Electronic Engineering. His main areas of interest are robotics and intelligent systems. His current research focuses on applications of imitation learning algorithms in autonomous driving systems. \end{IEEEbiography}

\begin{IEEEbiography}[{\includegraphics[width=1in,height=1.25in,clip,keepaspectratio]{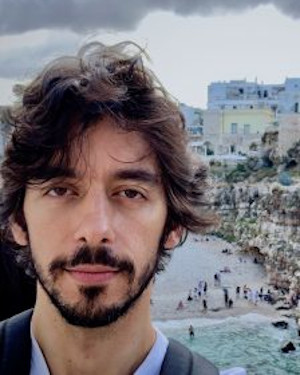}}]{Eric Aislan Antonelo}
received
the Ph.D. degree in Computer Engineering from Ghent University, Belgium, in 2011.
He is currently a Faculty Member with the Department of Automation and Systems Engineering, Federal University of Santa Catarina, Florianópolis, Brazil. His research interests include imitation learning for autonomous vehicles,
reservoir computing and machine learning for robotic or industrial applications. \end{IEEEbiography}

\end{document}